\documentclass[11pt]{article}

\usepackage[preprint]{acl}

\usepackage{times}
\usepackage{latexsym}

\usepackage[T1]{fontenc}
\usepackage[utf8]{inputenc}

\usepackage{microtype}

\usepackage{inconsolata}

\usepackage{graphicx}

\usepackage{amsmath}
\usepackage{amssymb}
\usepackage{algorithm}
\usepackage{algpseudocode}
\usepackage{comment}
\usepackage{booktabs}
\usepackage{array,tabularx}
\usepackage{placeins}
\usepackage{needspace}
\usepackage{xcolor}
\usepackage{listings}
\usepackage{tcolorbox}
\newcommand{\tablefont}{\fontsize{9}{10.5}\selectfont}
\newcommand{\thead}[1]{\begin{tabular}[c]{@{}c@{}}\bfseries #1\end{tabular}}

\title{D2: High-Capability Full-Duplex Spoken Language Models from Half-Duplex Backbones}

\author{
  \textbf{Cheng-Kuang Chang}$^{1,*}$,
  \textbf{Alexander H. Liu}$^{1}$,
  \textbf{Brian Kingsbury}$^{2}$, \\[0.15em]
  \textbf{Hilde Kuehne}$^{2}$,
  \textbf{Rogerio Feris}$^{2}$,
  \textbf{Samuel Thomas}$^{2}$, \\[0.15em]
  \textbf{James Glass}$^{1}$,
  \textbf{Kai-Wei Chang}$^{1,*}$ \\[0.5em]
  $^{1}$MIT CSAIL
  \qquad
  $^{2}$MIT-IBM Watson AI Lab \\
  $^{*}$\texttt{\{chang168, kwchang\}@mit.edu}
}

\begin{document}
\raggedbottom
\maketitle

\begin{abstract}
Full-duplex spoken language models (FD-SLMs) support simultaneous listening and speaking, but often lag behind half-duplex spoken language models (HD-SLMs) in general capabilities. We propose \textbf{Dialogue-to-Duplex (D2)}, a framework that converts pretrained HD-SLMs into high-capability FD-SLMs at fine temporal granularity. D2 adapts the original audio encoder to streaming through decoder-free representation distillation and learns full-duplex interaction through supervised fine-tuning. We systematically study the effects of chunk size on full-duplex interaction, general capabilities, and inference computation. We apply D2 to two distinct HD-SLM architectures, Qwen3-Omni and LLaMA-Omni 2. At a 12.5~Hz update rate, Qwen3-Omni-D2 achieves 99.2\% turn-taking success with 521-ms latency, 99.5\% interruption success, and a state-of-the-art VoiceBench score of 70.1.
\end{abstract}

\begin{comment}
1. interval -> chunk
2. k is chunk size 
kdelta chunk 
k chunk size
resolution
3. assistant -> model

Wording convention

\Delta encoder resolution
k chunk size
\Delta K chunk duration
[k model audio, k user audio, k autoregress] D2 chunk

k model audio = model audio chunk
k user audio = user audio chunk

bayling start HD-SLM -> FD-SLM good start but not yet comprehensizcely study blablalba, and bayling is dumb as fuck and we use D2 to sweep granulauryu and achieve SOTA.

\end{comment}

\begingroup
\renewcommand{\thefootnote}{}%
\footnotemark
\footnotetext{Demo: \url{https://d2-duplex-demo.github.io/}}
\addtocounter{footnote}{-1}%
\endgroup

\section{Introduction}
People continuously listen while speaking, interrupt each other, and respond without explicit turn boundaries~\citep{sacks1974systematic_turn_taking,levinson2015timing_in_turn_taking}. Modeling these dynamics remains a central challenge for spoken dialogue systems~\citep{skantze2021turn_taking_in_conversational_system}. Full-duplex spoken language models (FD-SLMs), including Moshi~\citep{defossez2024moshi} and MiniCPM-o 4.5~\citep{cui2026minicpm}, can perceive user audio while generating speech, supporting interruptions, overlapping speech, and low-latency turn taking.

\begin{figure}[t]
    \centering
    \includegraphics[width=\linewidth]
        {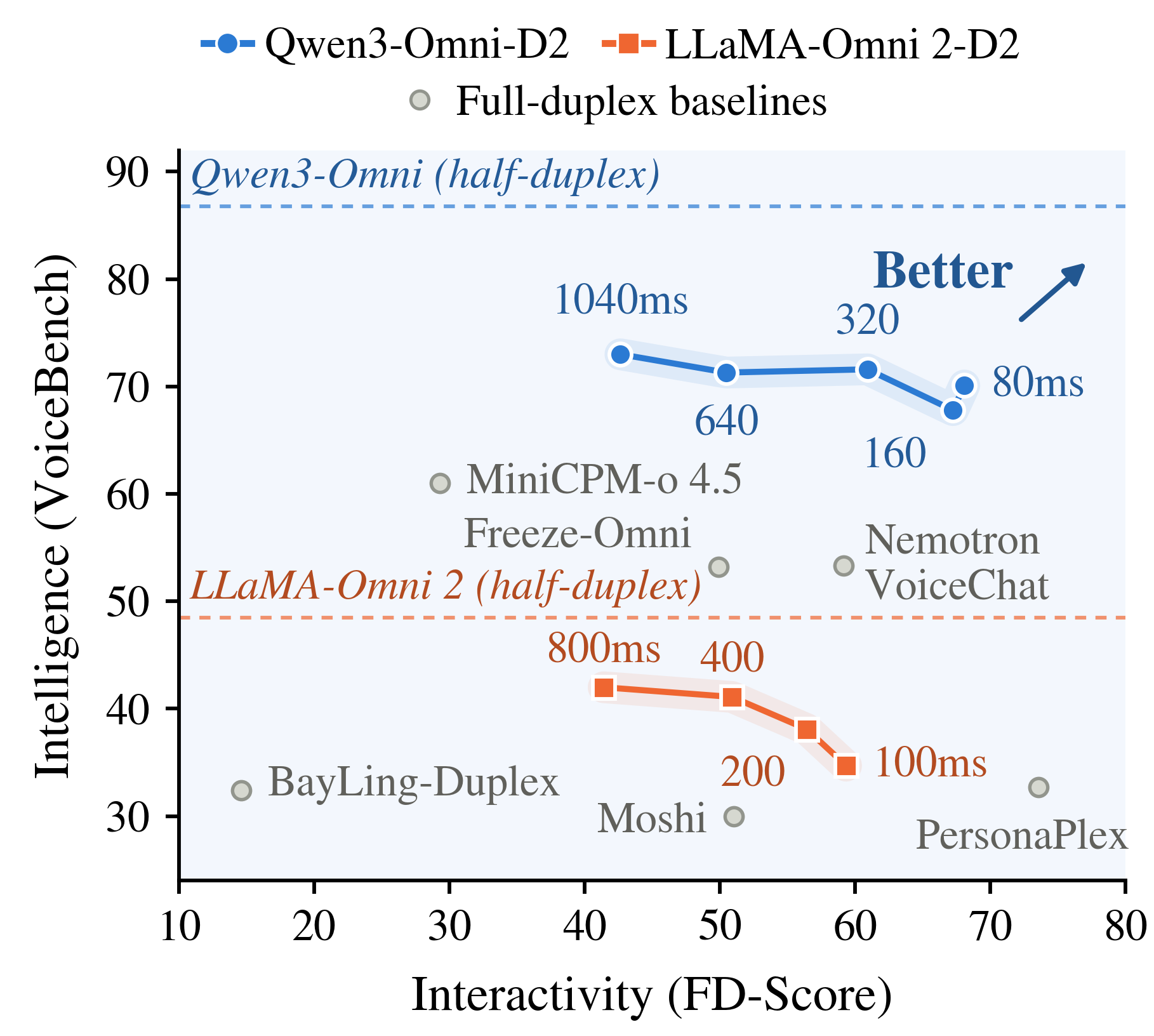}
    \caption{Capability and interactivity of full-duplex models.
    FD-Score is defined in
    Appendix~\ref{app:fd_definition}.}
    \label{fig:intro_fdscore}
\end{figure}

Despite these interaction capabilities, native FD-SLMs often lag behind strong half-duplex spoken language models (HD-SLMs) in audio understanding, knowledge, reasoning, and instruction following~\citep{chen2026voicebench}. HD-SLMs such as Qwen3-Omni~\citep{xu2025qwen3} and Kimi-Audio~\citep{ding2025kimi} provide stronger general capabilities. Converting these models to full-duplex operation offers a promising path toward high-capability FD-SLMs.

Previous work has explored converting HD-SLMs into FD-SLMs. BayLing-Duplex~\citep{fang2026bayling} converts GLM-4-Voice~\citep{zeng2024glm} to full-duplex operation, but its interactivity is limited by audio encoder blocks of up to 4~s and an 800-ms chunk duration. Moreover, increasing the update rate can lead to unstable generation and substantial performance degradation~\citep{cui2026minicpm}.

\begin{figure*}[t]
\centering
\includegraphics[width=\textwidth]{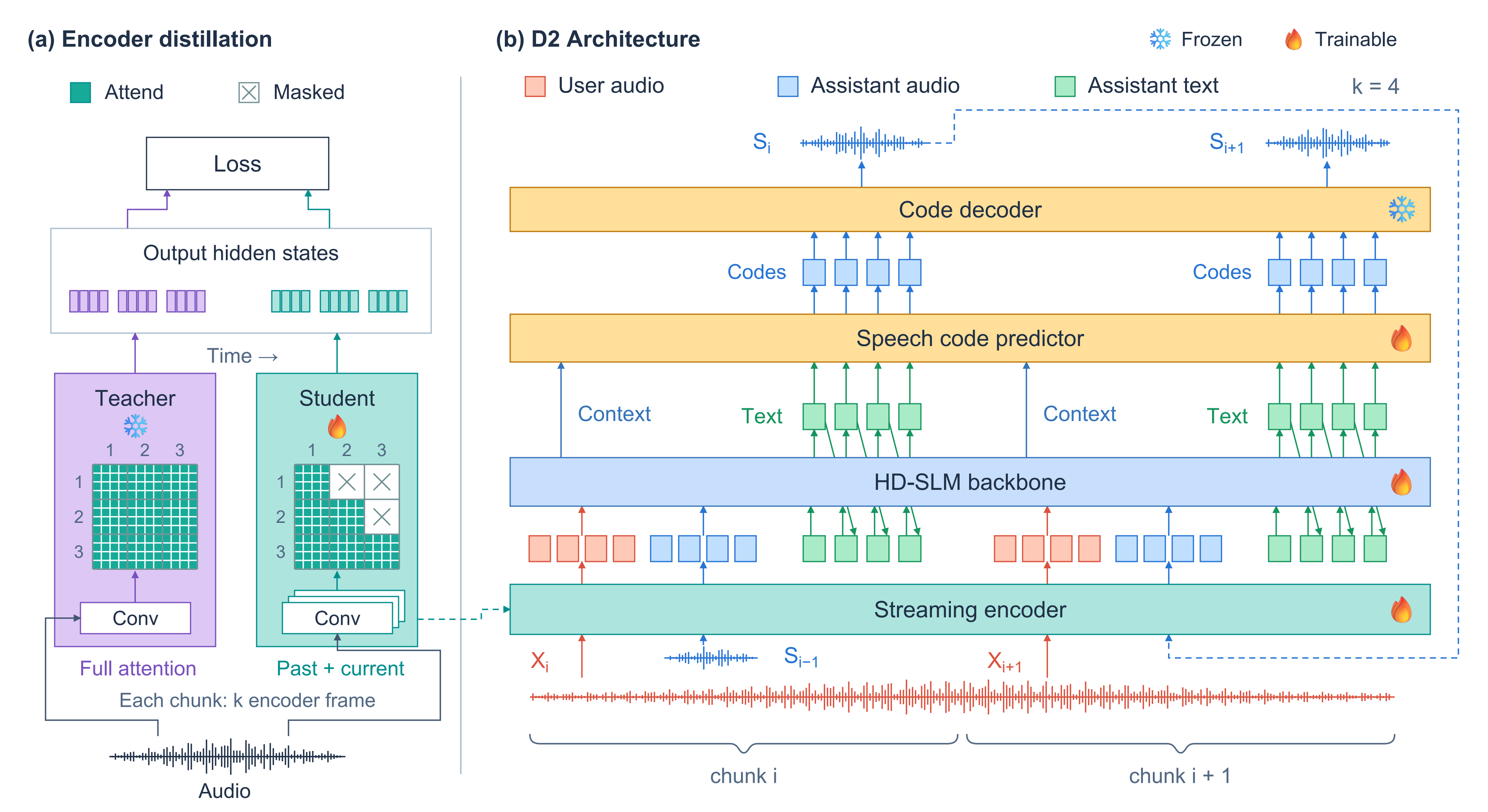}
\caption{\textbf{Overview of Dialogue-to-Duplex (D2).} (a) A frozen offline encoder supervises a streaming student with bidirectional attention within each chunk and causal attention across chunks. (b) At chunk $i$, D2 uses current user audio $X_i$ and preceding model audio $S_{i-1}$ to autoregressively generate $S_i$. Here, $k=4$.}
\label{fig:d2_architecture}
\end{figure*}

We propose \textbf{Dialogue-to-Duplex (D2)}, a framework for converting pretrained HD-SLMs into high-capability, fine-grained FD-SLMs (Figure~\ref{fig:d2_architecture}). D2 addresses the encoder streaming bottleneck through representation distillation and avoids the sharp performance degradation at higher update rates observed in prior work, maintaining strong general capabilities even at 12.5~Hz (80-ms chunks). We also systematically study the effects of chunk size on full-duplex interaction, general capabilities, and inference computation.

Applied to Qwen3-Omni, D2 achieves a state-of-the-art VoiceBench~\citep{chen2026voicebench} score of 70.1 at 12.5~Hz, outperforming the strongest evaluated full-duplex baseline by 9.1 points, while achieving 99.2\% turn-taking TOR and 99.5\% interruption TOR on Full-Duplex-Bench~\citep{lin2025fullduplexbench_v1}.

\Needspace{5\baselineskip}
In summary, our contributions are as follows:

\paragraph{High-capability HD-to-FD conversion.}
We convert pretrained HD-SLMs into FD-SLMs, achieving state-of-the-art VoiceBench performance among evaluated FD-SLMs and strong full-duplex interaction, while mitigating performance degradation at high update rates.

\paragraph{Decoder-free streaming adaptation.}
We reduce encoder latency through representation distillation without requiring the original training decoder or transcription labels.

\paragraph{Chunk-size trade-offs.}
We systematically study the effects of chunk size on full-duplex interaction, general capabilities, and inference computation.

\section{Related Work}

\subsection{Half-Duplex Spoken Language Models}
HD-SLMs combine speech understanding and generation with a language-model backbone. Earlier systems include SpeechGPT and AudioPaLM~\citep{zhang2023speechgpt,rubenstein2023audiopalm}, while VITA and Mini-Omni target low-latency or streaming speech interaction~\citep{fu2024vita,xie2024mini_omni}. Recent models such as GLM-4-Voice, LLaMA-Omni 2, Kimi-Audio, Qwen3-Omni, and Qwen3.5-Omni provide capable pretrained backbones for conversion to full-duplex operation~\citep{zeng2024glm,fang2025llama_omni_2,ding2025kimi,xu2025qwen3,qwen2026qwen3_5}.

\subsection{Full-Duplex Spoken Language Models}
Systems such as LSLM, Freeze-Omni, Moshi, and PersonaPlex support simultaneous listening and speaking, including interruption handling and overlapping speech~\citep{ma2025lslm,wang2024freeze,defossez2024moshi,roy2026personaplex}. MiniCPM-o 4.5 combines full-duplex interaction with stronger general capabilities, but uses a 1-s chunk duration and degrades substantially when this duration is reduced to 0.1~s~\citep{cui2026minicpm}. This motivates conversion methods that retain strong capabilities at smaller chunk durations.

\subsection{Conversion Strategy}
OmniFlatten converts a pretrained text LLM into a spoken-dialogue model through speech-modality alignment followed by full-duplex dialogue training~\citep{zhang2025omniflatten}. D2 starts from an HD-SLM that already possesses speech understanding and generation capabilities.

BayLing-Duplex~\citep{fang2026bayling} similarly converts an HD-SLM, GLM-4-Voice, to full-duplex operation. It retains a speech tokenizer requiring input blocks of up to 4~s and uses a large 800-ms chunk duration, resulting in latencies of several seconds. It also uses an additional DPO stage to stabilize turn-taking~\citep{rafailov2023dpo}. D2 adapts the original audio encoder to low-latency streaming through distillation and learns full-duplex interaction through supervised fine-tuning, supporting chunk durations as short as 80--100~ms.

\section{Method}

\subsection{Half-Duplex vs. Full-Duplex}
We formulate a spoken dialogue as a sequence of interaction units, $\mathcal{D}=\{(X_i,Y_i)\}_{i=1}^{N}$, where $X_i$ denotes the user audio observed in the $i$-th interaction unit and $Y_i$ denotes the corresponding model output, including speech and optional intermediate text tokens. Let $H_i=(X_{<i},Y_{<i})$ denote the preceding dialogue context. Given the user observations, the model factorizes the likelihood of its outputs as
\begin{equation}
p(Y_{1:N}\mid X_{1:N})=\prod_{i=1}^{N}p(Y_i\mid X_i,H_i).
\end{equation}

Half-duplex and full-duplex models differ primarily in how the interaction units are segmented. For a half-duplex model (HD-SLM), each $X_i$ corresponds to a complete user turn, yielding
\begin{equation}
p_{\mathrm{HD}}(Y\mid X)=\prod_{i=1}^{N_{\mathrm{turn}}}p(Y_i\mid X_i,H_i).
\end{equation}

A full-duplex model (FD-SLM), in contrast, operates at a finer temporal granularity. Let $\Delta$ denote the encoder resolution, i.e., the duration of one audio encoder output frame, and let $k\in\mathbb{N}_{>0}$ denote the chunk size, defined as the number of encoder frames processed per chunk. We partition the conversation into chunks of duration $k\Delta$, yielding
\begin{equation}
p_{\mathrm{FD}}(Y\mid X;k)=\prod_{i=1}^{N_k}p(Y_i\mid X_i,H_i),
\end{equation}

where each $X_i$ contains $k$ encoder frames and $N_k$ is the number of chunks. For example, Moshi~\citep{defossez2024moshi} uses $k=1$ and $\Delta=80$~ms, while MiniCPM-o 4.5~\citep{cui2026minicpm} uses $k=10$ and $\Delta=100$~ms. During each chunk, the model may generate speech or remain silent.

\subsection{Dialogue-to-Duplex (D2)}
The formulations above show that HD and FD models share a similar generation objective; they differ in the temporal granularity of the interaction sequence. An HD-SLM operates over variable-length, turn-defined units, whereas an FD-SLM operates over smaller chunks of duration $k\Delta$.

This suggests viewing full-duplex interaction as ordinary dialogue conducted at a finer temporal granularity. As $k$ decreases, the model observes and responds to the user more frequently, enabling increasingly fine-grained real-time interaction.

D2 converts pretrained HD-SLMs into FD-SLMs. Finer interaction granularity introduces two challenges. First, the audio encoder must operate with limited future context. Second, the model must learn to make conversational decisions at a finer temporal scale. D2 addresses these challenges through \textbf{streaming encoder distillation} followed by \textbf{D2 supervised fine-tuning}.

\subsection{Stage I: Encoder Distillation}
Audio encoders in pretrained HD-SLMs are typically non-streaming or blockwise streaming and rely on future acoustic context, making them unsuitable for low-latency full-duplex inference. Chunkwise attention and memory-based streaming encoders have been widely studied for low-latency speech recognition~\citep{shi2021emformer,yao2021wenet}. However, directly replacing the encoder with a streaming encoder may introduce a representation mismatch with the pretrained language model.

We instead convert the original encoder into a streaming student and distill it from the unchanged offline encoder~\citep{hinton2015distilling,chang2022distilhubert}. For each newly available chunk of duration $k\Delta$, the student recomputes the convolutional frontend using only the audio available up to the current chunk, while bidirectional self-attention in the Transformer is replaced with block-causal attention. Attention remains bidirectional within each chunk, but is causal across chunks. Given the same input audio, the original encoder serves as the teacher and provides frame-level target representations. We minimize:
\begin{equation}
\begin{aligned}
\mathcal{L}_{\mathrm{distill}}
={}& \frac{1}{TD}\sum_{t=1}^{T}\left\|\mathbf{h}^{\mathrm{stu}}_t-\mathbf{h}^{\mathrm{tea}}_t\right\|_1 \\
&-\frac{1}{T}\sum_{t=1}^{T}\log \sigma\!\left(\operatorname{cos}\!\left(\mathbf{h}^{\mathrm{stu}}_t,\mathbf{h}^{\mathrm{tea}}_t\right)\right).
\end{aligned}
\label{eq:encoder_distill}
\end{equation}
Here, $T$ and $D$ denote the sequence length and hidden dimension, respectively.

\subsection{Full-Duplex Inference}
D2 performs full-duplex inference in chunks of duration $k\Delta$. During each chunk, the model processes the newly accumulated user audio together with the self-generated audio from the preceding chunk, prefills their representations into the model context, and autoregressively generates the next $k$ output frames. User audio is captured continuously, including while the model is speaking. Algorithm~\ref{alg:d2_inference} summarizes the inference procedure.

\begin{algorithm}[!htbp]
\caption{D2 Full-Duplex Inference}
\label{alg:d2_inference}
\begin{algorithmic}[1]
\Require Streaming audio encoder $E$, dialogue model $M$, chunk duration $k\Delta$
\State Initialize model context $H$
\State $S_0 \gets \mathrm{Silence}(k\Delta)$
\For{each chunk $i=1,2,\ldots$}
    \State Observe user audio $X_i$ of duration $k\Delta$
    \State $U_i \gets E(X_i)$ \Comment{$k$ user embeddings}
    \State $A_i \gets E(S_{i-1})$ \Comment{$k$ self embeddings}
    \State $H \gets \mathrm{Prefill}_M(H,[U_i;A_i])$
    \State $Y_i \gets \emptyset$
    \For{$j=1,\ldots,k$}
        \State $(y_{i,j},H) \gets \mathrm{GenerateFrame}_M(H)$
        \State $Y_i \gets Y_i \mathbin{\|} y_{i,j}$
    \EndFor
    \State $S_i \gets \mathrm{Render}(Y_i)$
    \State Emit $S_i$
\EndFor
\end{algorithmic}
\end{algorithm}

Here, $y_{i,j}$ denotes an output frame containing speech together with its associated intermediate text. The preceding output $S_{i-1}$ provides delayed self-audio feedback.

Increasing $k$ results in coarser interaction granularity and weaker full-duplex responsiveness. Under our formulation, an HD-SLM lies at the coarse-grained extreme of this spectrum, where interaction boundaries are determined by complete conversational turns rather than short chunks.

\subsection{Stage II: D2 SFT}
We train the model to follow the inference procedure in Algorithm~\ref{alg:d2_inference} by converting ordinary dialogues into temporally aligned sequences of fixed-duration chunks. User and assistant audio are placed on a shared timeline that preserves silence and overlapping speech, and the timeline is partitioned into chunks of duration $k\Delta$. For each chunk $i$, $X_i$ contains the newly observed user audio, $S_{i-1}$ contains the assistant audio from the preceding chunk, and $Y_i$ contains the target $k$ output frames. Explicit control tokens indicate response initiation and interruption, while padding tokens occupy positions without lexical output.

Starting from the pretrained HD-SLM equipped with the streaming encoder from Stage I, we fine-tune the model with teacher forcing using the same temporal ordering as during inference. Let $O_i=[E(X_i);E(S_{i-1})]$ denote the audio observations prefilled at chunk $i$. We minimize
\begin{equation}
\begin{aligned}
\mathcal{L}_{\mathrm{D2}}
&=-\mathbb{E}_{\mathcal{D}}\!\left[\sum_{i=1}^{N_k}\sum_{j=1}^{k} w_{i,j}\right.\\
&\qquad\left.{}\cdot\log p_{\theta}\!\left(y_{i,j}\mid O_i,H_i,Y_{i,<j}\right)\right].
\end{aligned}
\label{eq:d2_sft}
\end{equation}
where $w_{i,j}$ controls output weighting and masking. We upweight sparse conversational control outputs, such as response-initiation and interruption events, while downweighting padding outputs. This trains the HD-SLM to follow the same fine-grained inference procedure used by D2.

\section{Experimental Setup}
We describe the HD-SLM backbones, datasets, training procedure, and evaluation benchmarks used for D2.

\subsection{Half-Duplex Models}
We apply D2 to Qwen3-Omni and LLaMA-Omni 2~\citep{xu2025qwen3,fang2025llama_omni_2}, whose streaming speech decoders support low-latency output. The temporal resolution of their audio representations determines $\Delta$ used in D2.

\paragraph{Qwen3-Omni.}
Qwen3-Omni-30B-A3B uses a Mixture-of-Experts Thinker--Talker architecture with an Audio Transformer (AuT) encoder, a vision encoder, and a streaming codec decoder. D2 uses only its audio input pathway. The AuT convolutional frontend downsamples the input to 12.5~Hz ($\Delta=80$~ms). Its Transformer attends bidirectionally within audio blocks; Stage I adapts this encoder to streaming. The Thinker generates text, while the Talker produces 16-codebook speech frames conditioned on text embeddings and Thinker hidden states. Code2Wav incrementally converts these frames into waveforms.

\paragraph{LLaMA-Omni 2.}
We use the bilingual 7B variant, built on Qwen2.5. Its Whisper-large-v3 encoder feeds a speech adapter that concatenates five consecutive encoder frames and projects them into the language model's embedding space. The resulting representations have a rate of 10~Hz ($\Delta=100$~ms). A separate autoregressive TTS model generates speech conditioned on text-token embeddings and language-model hidden states, supporting streaming speech output.

\subsection{Datasets}

\paragraph{Stage I: Encoder distillation.}
We use the LibriSpeech \texttt{train-\allowbreak clean-\allowbreak 100} and \texttt{train-\allowbreak clean-\allowbreak 360} splits~\citep{panayotov2015librispeech} as speech data, together with the audio tracks from AVQA~\citep{yang2022avqa} as general audio data. AVQA provides non-speech events and environmental sounds, such as rainfall and traffic.

\begin{figure}[t]
  \centering
  \includegraphics[width=\columnwidth]{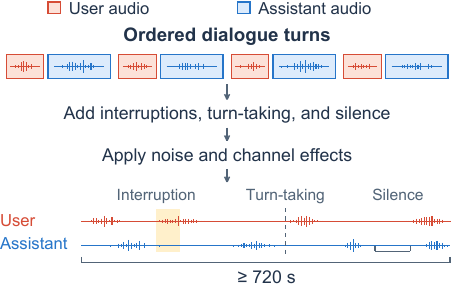}
  \caption{Full-duplex data construction. Ordered turns become sequences of at least 720~s with interruptions, turn-taking, and silence.}
  \label{fig:data_concat}
\end{figure}

\paragraph{Stage II: D2 training.}
Figure~\ref{fig:data_concat} illustrates how we construct full-duplex training sequences from the turn-based conversations in InstructS2S-\allowbreak 200K~\citep{fang2025llama_omni} and SODA~\citep{kim2023soda}. We also include AVQA to preserve general audio understanding. We resynthesize model responses using Kokoro TTS~\citep{hexgrad2025kokoro} to provide a consistent output voice.

To construct each training sequence, we concatenate complete user--model turn pairs in their original order until the sequence spans at least 720~s of audio. At each transition to the next user turn, we sample one of three transition types: interruption with 50\% probability, direct concatenation with 30\%, or a silent gap with 20\%. An \texttt{INTERRUPT} target placed after user-speech onset teaches the model to stop its current response, while a \texttt{RESPONSE} target placed after user-speech offset indicates that a new response should begin. An audio \texttt{EOS} target terminates model speech either at its natural end or upon interruption.

Additive environmental noise and channel distortions augment half of the constructed InstructS2S-\allowbreak 200K and SODA sequences. AVQA audio remains unaugmented.

\subsection{Implementation Details}

\paragraph{Stage I: Encoder distillation.}
We train each streaming audio encoder for 30,000 updates with a batch size of four, corresponding to approximately 338 hours of LibriSpeech and 66.4 hours of AVQA. We use constant learning rates of $5\times10^{-5}$ for the convolutional frontend and $2\times10^{-6}$ for the fully fine-tuned attention layers. The remaining encoder components use LoRA~\citep{hu2021lora} with a learning rate of $1\times10^{-4}$.

\paragraph{Stage II: D2 training.}
We train each D2 model for 10,000 updates with a batch size of four, corresponding to approximately 380 hours of InstructS2S-\allowbreak 200K, 200 hours of SODA, and 100 hours of AVQA. We use Transformer-XL-style training~\citep{dai2019transformerxl} with 90-s training segments and a detached 90-s KV cache carried over from the preceding segment.

Text cross-entropy weights are 1 for lexical tokens, 0.05 for \texttt{PAD}, and 20 for \texttt{RESPONSE} and \texttt{INTERRUPT}. Ordinary speech-code targets have weight 1, while audio \texttt{EOS} has weight 20 for Qwen3-Omni and 3 for LLaMA-Omni 2. Trainable modules, LoRA configurations, learning rates, schedules, and hardware are detailed in Appendix~\ref{app:training}.

\begin{table*}[t]
\centering
\tablefont
\setlength{\tabcolsep}{3pt}
\begin{tabular*}{\textwidth}{@{\extracolsep{\fill}}lrrrrrrr@{}}
\toprule
\thead{Model} & \thead{Chunk duration\\(ms)} & \thead{Quality\\(1--5) $\uparrow$} & \thead{Turn taking\\(\%) $\uparrow$} & \thead{TT latency\\(s) $\downarrow$} & \thead{Interrupt\\(\%) $\uparrow$} & \thead{Int. latency\\(s) $\downarrow$} & \thead{Pause\\(\%) $\uparrow$} \\
\midrule
Qwen3-Omni-D2 & 80 & 4.56 & 99.2 & 0.521 & 99.5 & 0.403 & 21.3 \\
 & 160 & 4.67 & 100 & 0.539 & 99.5 & 0.432 & 19.0 \\
 & 320 & 4.61 & 100 & 0.710 & 99.5 & 0.571 & 25.5 \\
 & 640 & 4.62 & 99.2 & 1.01 & 99.5 & 0.928 & 55.1 \\
 & 1040 & 4.84 & 100 & 1.38 & 99.5 & 1.30 & 61.6 \\
\midrule
LLaMA-Omni 2-D2 & 100 & 3.33 & 100 & 0.746 & 99.5 & 0.621 & 74.5 \\
 & 200 & 3.62 & 99.2 & 0.825 & 100 & 0.710 & 78.2 \\
 & 400 & 3.69 & 100 & 1.02 & 99.0 & 0.894 & 69.4 \\
 & 800 & 4.03 & 98.3 & 1.45 & 99.0 & 1.32 & 70.8 \\
\midrule
BayLing-Duplex$^{\dagger}$ & -- & 3.55 & 85.7 & 5.57 & 100 & 5.19 & 72.2 \\
Freeze-Omni & -- & 3.82 & 34.5 & 0.288 & 98.5 & 0.349 & 69.4 \\
MiniCPM-o 4.5 & -- & 4.45 & 75.6 & 1.92 & 94.5 & 1.89 & 91.2 \\
PersonaPlex & -- & 4.43 & 99.2 & 0.346 & 95.0 & 0.294 & 31.9 \\
Moshi & -- & 3.40 & 100 & 0.615 & 89.5 & 1.23 & 0.0 \\
Nemotron VoiceChat & -- & 3.97 & 91.6 & 0.584 & 99.0 & 0.638 & 60.2 \\
\bottomrule
\end{tabular*}
\caption{Full-Duplex-Bench results. Chunk duration is $k\Delta$ for D2 models; -- marks non-D2 models. Quality is post-interruption response quality rated by GPT-4o. Turn taking/Interrupt are take-over rates (TOR); Pause is $100$ minus Synthetic Pause Handling TOR (\%). TT/Int. mean turn-taking/interruption.}

\label{tab:fdb_results}
\end{table*}

\subsection{Benchmarks}

\paragraph{Full-Duplex-Bench.}
Full-Duplex-Bench evaluates interactive full-duplex behavior~\citep{lin2025fullduplexbench_v1}. We focus on three dimensions. \textit{Turn taking} measures whether the model begins responding after the user finishes speaking. \textit{Interruption handling} measures whether the model yields when interrupted and whether it can generate an appropriate response after the user's turn. \textit{Pause handling} measures whether the model avoids falsely taking the turn during short pauses within a user's utterance.

\paragraph{VoiceBench.}
VoiceBench~\citep{chen2026voicebench} evaluates general spoken-language capabilities through nine sub-benchmarks spanning question answering, reasoning, instruction following, and safety. AlpacaEval~\citep{li2023alpacaeval}, CommonEval~\citep{chen2026voicebench,ardila2020common_voice}, and WildVoice~\citep{chen2026voicebench} evaluate open-ended question answering. SD-QA~\citep{faisal2021sdqa} evaluates reference-based question answering, while OpenBookQA (OBQA)~\citep{mihaylov2018open_book_qa} and MMSU~\citep{chen2026voicebench,wang2024mmlu} evaluate multiple-choice question answering. BBH~\citep{suzgun2023bbh} evaluates reasoning, IFEval~\citep{zhou2023ifeval} evaluates instruction following, and AdvBench~\citep{zou2023adv_bench} evaluates safety. We evaluate the first 200 samples from each sub-benchmark to assess capability retention relative to the original HD-SLMs and compare D2 with other FD-SLMs.

\begingroup
\renewcommand{\thefootnote}{\fnsymbol{footnote}}
\footnotetext[2]{BayLing-Duplex results were obtained using
the authors' official implementation under our evaluation protocol.}
\endgroup
\begin{table*}[t]
\centering
\tablefont
\setlength{\tabcolsep}{2pt}
\begin{tabular*}{\textwidth}{@{\extracolsep{\fill}}lrrrrrrrrrrr@{}}
\toprule
\thead{Model} & \thead{Chunk\\duration\\(ms)} & \thead{Alpaca\\Eval} & \thead{Common\\Eval} & \thead{Wild\\Voice} & \thead{SD-\\QA} & \thead{MMSU} & \thead{OBQA} & \thead{BBH} & \thead{IF\\Eval} & \thead{Adv\\Bench} & \thead{Overall} \\
\midrule
Qwen3-Omni-D2 & 80 & 4.30 & 3.47 & 3.58 & 64.5 & 64.5 & 81.5 & 64.5 & 37.2 & 91.5 & 70.1 \\
 & 160 & 4.33 & 3.44 & 3.63 & 68.0 & 60.0 & 71.5 & 59.0 & 29.6 & 94.0 & 67.8 \\
 & 320 & 4.40 & 3.70 & 3.76 & 67.0 & 60.0 & 86.5 & 62.0 & 36.9 & 95.0 & 71.6 \\
 & 640 & 4.38 & 3.96 & 3.78 & 57.0 & 63.0 & 85.5 & 60.0 & 38.0 & 95.5 & 71.3 \\
 & 1040 & 4.53 & 3.67 & 3.94 & 66.5 & 63.5 & 86.0 & 60.0 & 39.4 & 98.5 & 73.0 \\
Qwen3-Omni & HD & 4.84 & 4.52 & 4.69 & 68.5 & 73.0 & 91.0 & 87.5 & 80.7 & 99.5 & 86.8 \\
\midrule
LLaMA-Omni 2-D2 & 100 & 3.38 & 1.94 & 2.13 & 9.0 & 27.5 & 36.0 & 48.5 & 12.5 & 29.5 & 34.7 \\
 & 200 & 3.59 & 2.31 & 2.28 & 15.0 & 30.5 & 37.0 & 50.5 & 11.7 & 35.0 & 38.1 \\
 & 400 & 3.52 & 2.48 & 2.35 & 18.5 & 26.5 & 43.0 & 51.5 & 13.5 & 49.5 & 41.1 \\
 & 800 & 3.59 & 2.86 & 2.64 & 20.5 & 32.0 & 40.0 & 47.5 & 15.8 & 40.5 & 42.0 \\
LLaMA-Omni 2 & HD & 3.94 & 3.50 & 3.25 & 29.0 & 25.0 & 44.5 & 48.5 & 14.6 & 61.5 & 48.5 \\
\midrule
Moshi & -- & 2.04 & 1.53 & 1.49 & 17.5 & 22.5 & 21.0 & 47.0 & 13.2 & 47.5 & 30.0 \\
PersonaPlex & -- & 3.14 & 2.30 & 2.31 & 21.0 & 27.5 & 24.0 & 49.5 & 12.5 & 5.0 & 32.7 \\
BayLing-Duplex & -- & 3.24 & 2.23 & 2.13 & 6.0 & 30.5 & 28.0 & 49.5 & 14.2 & 11.0 & 32.4 \\
Freeze-Omni & -- & 3.74 & 3.47 & 3.20 & 45.0 & 26.0 & 33.5 & 47.0 & 21.5 & 98.0 & 53.2 \\
MiniCPM-o 4.5 & -- & 3.57 & 3.28 & 3.02 & 51.0 & 46.0 & 72.5 & 57.5 & 37.9 & 86.5 & 61.0 \\
Nemotron VoiceChat & -- & 3.48 & 2.79 & 2.83 & 28.5 & 36.5 & 60.0 & 52.5 & 20.3 & 100 & 53.3 \\
Kimi-Audio & HD & 4.15 & 3.06 & 3.88 & 55.0 & 57.0 & 86.5 & 73.0 & 57.9 & 99.5 & 72.3 \\
\bottomrule
\end{tabular*}
\caption{VoiceBench scores. Chunk duration is $k\Delta$ for D2 models; HD denotes a half-duplex model and -- denotes another full-duplex model. AlpacaEval, CommonEval, and WildVoice use a 1--5 scale. The remaining sub-benchmarks report percentages; Overall is the aggregate score on a 0--100 scale.}
\label{tab:voicebench}
\end{table*}

\begin{figure*}[t]
  \centering
  \includegraphics[width=\textwidth]{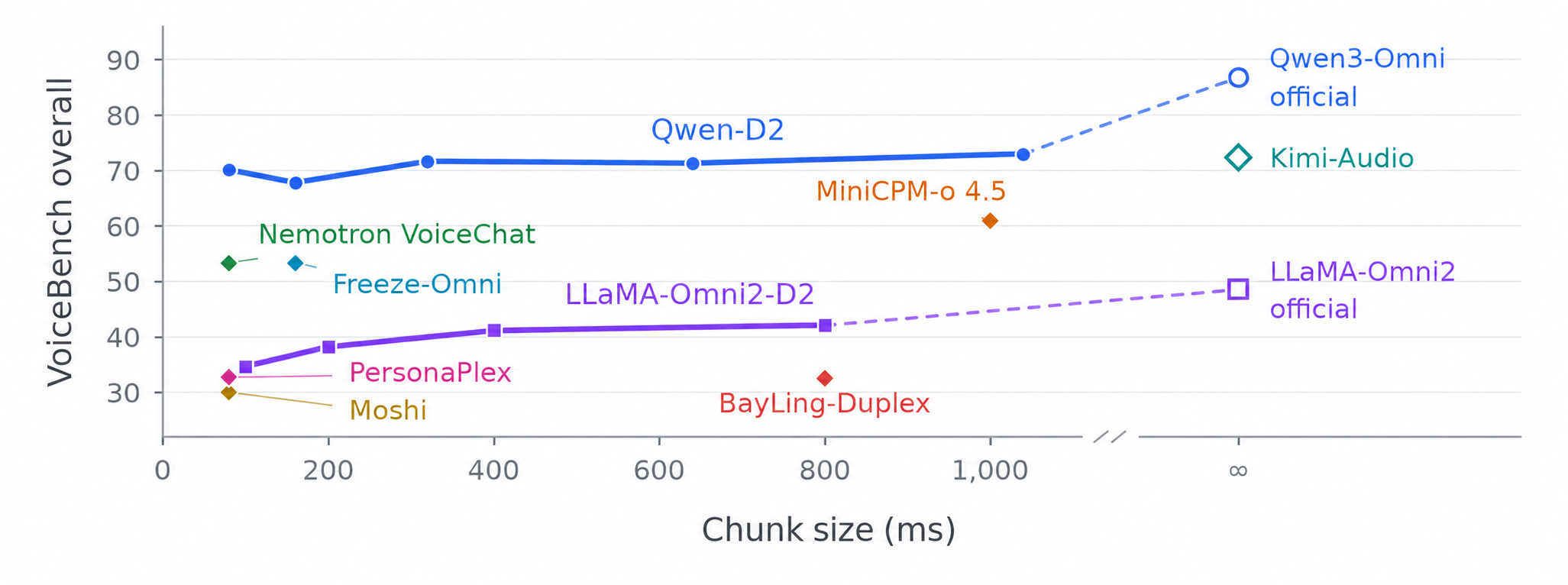}
  \caption{VoiceBench versus chunk duration. Filled circles/squares and solid lines show D2 variants; open markers show original half-duplex models. Dashed connections compare D2 with its backbones. The axis break and $\infty$ denote turn-based operation without fixed-duration chunks.}
  \label{fig:voicebench_landscape}
\end{figure*}

\section{Results}
\subsection{Full-Duplex-Bench}
We compare D2 with six full-duplex baselines in Table~\ref{tab:fdb_results}~\citep{fang2026bayling,wang2024freeze,cui2026minicpm,roy2026personaplex,defossez2024moshi,nvidia2026voicechat}. Both Qwen3-Omni and LLaMA-Omni 2 achieve strong full-duplex performance after D2 conversion. Turn-taking and interruption take-over rates (TOR) remain close to 100\% across all chunk sizes. Qwen3-Omni-D2 also achieves the highest post-interruption response quality among the evaluated models, with GPT scores of 4.56--4.84.

At the smallest chunk size, Qwen3-Omni-D2 achieves turn-taking and interruption latencies of 521~ms and 403~ms, respectively, while LLaMA-Omni 2-D2 achieves 746~ms and 621~ms. Larger chunk sizes increase latency but generally improve post-interruption response quality. Qwen3-Omni-D2's pause-handling success increases from 21.3\% at 80~ms to 61.6\% at 1040~ms, showing the tradeoff between responsiveness and tolerance of user pauses.

\subsection{VoiceBench}
We compare D2 with its original backbones and seven spoken-language baselines in Table~\ref{tab:voicebench}~\citep{xu2025qwen3,fang2025llama_omni_2,defossez2024moshi,roy2026personaplex,fang2026bayling,wang2024freeze,cui2026minicpm,nvidia2026voicechat,ding2025kimi}. Figure~\ref{fig:voicebench_landscape} shows the results across chunk durations alongside the original half-duplex backbones.

Qwen3-Omni-D2 achieves the highest overall VoiceBench score among the evaluated full-duplex models at every tested chunk duration. Even at 80~ms, it scores 70.1, exceeding the strongest baseline, MiniCPM-o 4.5, by about 9.1 points. At 1040~ms, it reaches 73.0, comparable to the half-duplex model Kimi-Audio's 72.3. These results demonstrate that D2 brings strong general capabilities to full-duplex operation while supporting fine-grained interaction.

\subsection{Computation Analysis}
We measure the full inference computation time of Qwen3-Omni-D2 on a single RTX PRO 6000 GPU (96 GB), using BF16 and a batch size of one. Table~\ref{tab:computation_time} shows that both mean and p95 computation times remain below the chunk duration at all tested chunk sizes, supporting real-time inference under this setting. Larger chunk sizes yield lower real-time factors by amortizing the audio encoder's computation over larger input chunks. Measurement details are provided in Appendix~\ref{app:computation}.

\begin{table}[t]
\centering
\tablefont
\begin{tabular*}{\columnwidth}
{@{\extracolsep{\fill}}cccc@{}}
\toprule
\thead{Chunk\\duration (ms)} & \textbf{Mean (ms)} & \textbf{p95 (ms)} & RTF $\downarrow$ \\
\midrule
80    & 56.9 & 57.2 & 0.711 \\
160   & 110  & 110  & 0.688 \\
320   & 216  & 217  & 0.676 \\
640   & 429  & 431  & 0.670 \\
1,040 & 682  & 684  & 0.656 \\
\bottomrule
\end{tabular*}
\caption{Full inference computation time of Qwen3-Omni-D2
with a fixed 2,048-slot Thinker KV cache per layer.
RTF is computation time divided by chunk duration.}
\label{tab:computation_time}
\end{table}

\section{Ablation}

\begin{table}[t]
\centering
\tablefont
\setlength{\tabcolsep}{3pt}
\begin{tabular*}{\columnwidth}{@{\extracolsep{\fill}}lrrr@{}}
\toprule
\thead{Metric} & \thead{Distill.}
& \thead{Supervised} & \thead{No Stage I} \\
\midrule
Pause (\%) $\uparrow$    & 74.5  & 61.6  & 62.0 \\
Int. TOR (\%) $\uparrow$   & 99.5  & 100 & 99.0 \\
TT TOR (\%) $\uparrow$ & 100 & 100 & 100 \\
Quality (1--5) $\uparrow$       & 3.33   & 3.09   & 2.75 \\
Int. latency (ms) $\downarrow$  & 621    & 590    & 617 \\
TT latency (ms) $\downarrow$    & 746    & 676    & 686 \\
VoiceBench $\uparrow$  & 34.7  & 35.4  & 30.9 \\
\bottomrule
\end{tabular*}
\caption{Stage-I encoder training ablation for LLaMA-Omni 2-D2 with a 100-ms chunk duration. Distill. denotes distillation; No Stage I omits encoder adaptation.}
\label{tab:encoder_distill_vs_supervised}
\end{table}

\subsection{Stage-I Encoder Distillation}

Before Stage II, we adapt the audio encoder to streaming through distillation. We examine whether Stage I improves the converted model and whether supervised encoder training can serve as an alternative. Distillation requires only the pretrained encoder, allowing D2 conversion even when its training decoder is not publicly available. When the decoder is available, Stage I can instead use the original speech-recognition objective.

We evaluate both approaches and a baseline without Stage I using LLaMA-Omni 2-D2 with a 100-ms chunk duration. Table~\ref{tab:encoder_distill_vs_supervised} shows that omitting Stage I lowers VoiceBench from 34.7 to 30.9 and post-interruption quality from 3.33 to 2.75. Stage I therefore improves response quality. Supervised fine-tuning achieves comparable results, with a slightly higher VoiceBench score but lower post-interruption quality than distillation.

\subsection{Control Signal}
\label{sec:control-signal-ablation}

In Stage II, we upweight control signal tokens and downweight PAD tokens in the text cross-entropy loss to improve turn control and reduce under-emission. We examine these choices using LLaMA-Omni 2-D2 with a 100-ms chunk duration. We fine-tune the model for 1,000 additional updates under each weight configuration, and evaluate all four models on Full-Duplex-Bench.

\begin{table}[t]
\centering
\tablefont
\setlength{\tabcolsep}{3pt}
\begin{tabular*}{\columnwidth}{@{\extracolsep{\fill}}rrrrr@{}}
\toprule
\multicolumn{2}{c}{CE weight}
& Turn taking
& \multicolumn{2}{c}{Interruption} \\
\cmidrule(lr){1-2}\cmidrule(lr){3-3}\cmidrule(lr){4-5}
PAD & Control
& TOR $\uparrow$
& TOR $\uparrow$ & Score $\uparrow$ \\
\midrule
0.05 & 20 & \textbf{100} & \textbf{99.0} & \textbf{3.67} \\
1    & 20 & 95.8 & 98.0 & 3.44 \\
0.05 & 1  & 95.8 & 98.5 & 3.55 \\
1    & 1  & 62.2 & 90.5 & 3.45 \\
\bottomrule
\end{tabular*}
\caption{Control-signal weighting ablation on
Full-Duplex-Bench. Lexical-token weight is 1 throughout.
TOR is reported in percentages. Scores cover qualifying
interruption responses.}
\label{contro_signal_abla}
\end{table}

Table~\ref{contro_signal_abla} shows that removing either weighting adjustment lowers turn-taking TOR from 100\% to 95.8\%. Removing both reduces turn-taking and interruption TOR to 62.2\% and 90.5\%, respectively. The original weighting also gives the highest post-interruption score. Both adjustments help maintain response initiation and turn control.

\FloatBarrier
\section{Conclusion}
We proposed \textbf{Dialogue-to-Duplex (D2)}, a framework for converting HD-SLMs into FD-SLMs through streaming encoder distillation and full-duplex supervised fine-tuning, using several hundred hours of training data per stage. D2 enables low-latency streaming and strong full-duplex interaction while retaining substantial general capabilities of the original HD-SLM. Across two different HD-SLM architectures, D2 achieves strong results on full-duplex interaction benchmarks and state-of-the-art VoiceBench performance among evaluated full-duplex models, narrowing the capability gap between FD-SLMs and HD-SLMs.

\section*{Limitations}
First, the speech synthesis module contributes a large portion of the end-to-end latency in D2. Reducing speech-generation latency often leads to degraded speech quality or unstable generation, which restricts the choice of HD-SLM backbones to models with sufficiently low-latency speech synthesis. Second, we construct full-duplex training data from turn-based conversations because high-quality publicly available full-duplex dialogue data is currently limited. Finally, pause handling remains limited, particularly for Qwen3-Omni-D2 at small chunk durations.

\section*{Ethical Considerations}
Full-duplex spoken language models raise privacy and
consent concerns because they continuously process
incoming audio, including while generating speech.
Their natural, low-latency interactions may also
facilitate impersonation, social engineering, and
deceptive conversations. Furthermore, D2 may inherit
biases and harmful behaviors from pretrained models.
Responsible deployment should therefore consider
informed user consent, transparency regarding
AI-generated speech, privacy-preserving audio
processing, and safeguards against misuse.


\bibliography{custom}

\clearpage
\appendix

% Original full-width appendix algorithm boxes.
\floatstyle{plain}
\restylefloat{algorithm}
\newtcolorbox{dTwoCodeBox}[1]{
  colback=blue!3,
  colframe=blue!50!black,
  colbacktitle=blue!50!black,
  coltitle=white, fonttitle=\bfseries,
  title={#1}, sharp corners=south, boxrule=0.6pt,
  left=6pt, right=6pt, top=4pt, bottom=4pt,
  before skip=0pt, after skip=0pt
}
\lstdefinestyle{d2code}{
  language=Python, basicstyle=\ttfamily\fontsize{10}{12}\selectfont,
  keywordstyle=\bfseries, commentstyle=,
  columns=fullflexible, keepspaces=true, showstringspaces=false,
  breaklines=true, aboveskip=3pt, belowskip=3pt,
  frame=none
}

\section{Real-Time Computation}
\label{app:computation}
\paragraph{Hardware and inputs.}
We use one NVIDIA RTX PRO 6000 Blackwell Server Edition GPU (96 GB),
one CPU core, and a batch size of one. Neural weights and activations use
BF16; the CPU log-Mel frontend uses FP32. The benchmark uses synthetic PCM
inputs. We use PyTorch 2.11.0+cu128, Transformers 5.13.0, SDPA attention,
\texttt{batched\_mm} experts, and CUDA graphs.

\paragraph{Timed operations.}
Timing runs from CPU input PCM to CPU output PCM. It includes
waveform-to-Mel conversion, both streaming encoders for user and delayed
self audio, Thinker prefill and decoding, text-token selection, Talker,
all 15 residual-code predictions per speech frame, and incremental
Code2Wav decoding. Data transfers, codec sampling, and PCM conversion
are included. Each chunk generates $k$ text tokens and $k$ speech-code
frames, with codec EOS disabled to measure continuous speech generation.
Initialization, graph capture, cache reset, audio acquisition, transport,
and playback are outside the timed region.

\paragraph{Measurement protocol.}
We evaluate $k\in\{1,2,4,8,13\}$. The CUDA graphs preallocate 2,048
Thinker KV-cache slots per layer for every $k$; unwritten slots are masked.
Each measurement starts with zero conversational history and one Talker
speaker-prefix token. We report the mean and 95th percentile. RTF is
$t/(k\Delta)$, where $t$ is computation time and $\Delta=80$ ms.

\section{Implementation Pseudocode}
\label{app:pseudocode}

\subsection{Stage-I Streaming Encoder}

\paragraph{Qwen3-Omni.}
Each chunk supplies $8k$ new Mel frames at 100 Hz. We run the
convolutional frontend once on the latest 104 Mel frames, then keep
the last $k$ of its 13 output positions. Each Transformer layer
attends to these $k$ positions and at most $104-k$ cached positions.
Attention is bidirectional within the current chunk; future chunks
are absent. User and self audio have separate states but share
encoder weights. Mel history and per-layer KV caches are initially
empty. Algorithm~\ref{alg:app-qwen-encoder} gives one streaming step.

\begin{algorithm*}[!t]
\refstepcounter{algorithm}\label{alg:app-qwen-encoder}
\begin{dTwoCodeBox}{Algorithm~\thealgorithm: Stage-I streaming encoder for Qwen3-Omni.}
\begin{lstlisting}[style=d2code]
def encoder_step(new_mel, state, k):
    # new_mel: [B, 128, 8*k]
    w = cat([state.mel, new_mel], dim=-1)
    w = left_pad(w[..., -104:], length=104)
    x = w[:, None]  # [B,1,128,104]
    for conv in (conv1, conv2, conv3):
        x = gelu(conv(x))
    x = conv_out(flatten_channels(x))
    h = x[:, -k:] + position[13 - k : 13]
    # h: [B,k,1280]; Q/K/V split into heads
    for layer, block in enumerate(blocks):
        q, key, val = block.qkv(block.norm1(h))
        key = cat([state.K[layer], key], dim=-2)
        val = cat([state.V[layer], val], dim=-2)
        a = sdpa(q, key, val, is_causal=False)
        h = h + block.out(merge_heads(a))
        h = h + block.mlp(block.norm2(h))
        state.K[layer] = key[..., -(104 - k) :, :]
        state.V[layer] = val[..., -(104 - k) :, :]
    state.mel = w
    return proj2(gelu(proj1(norm(h)))), state
    # output: [B,k,2048]
\end{lstlisting}
\end{dTwoCodeBox}
\end{algorithm*}

For Stage-I training, we batch the convolution windows and apply
the same attention pattern to the resulting sequence. For query
position $q$, let $b=k\lfloor q/k\rfloor$. Key position $p$ is visible
exactly when $\max(0,b-(104-k))\leq p<b+k$.
We match the student output to the frozen original encoder using
the Stage-I loss. Gradients pass through all valid positions in the
training sequence; the streaming cache in Algorithm~\ref{alg:app-qwen-encoder}
describes inference.

\paragraph{LLaMA-Omni 2.}
Whisper receives $10k$ new Mel frames and produces $5k$ encoder rows
per chunk. Its convolution uses the last two Mel frames plus the
current input. Five consecutive Transformer rows are concatenated
and projected to one 3,584-dimensional embedding. The implementation
retains convolution outputs and recomputes block-causal attention
within a window of $R=5k\lfloor1500/(5k)\rfloor$ rows; it does not
use the Qwen encoder's 104-token KV cache.
The initial two-frame Mel tail is zero.
Algorithm~\ref{alg:app-whisper-encoder} gives the streaming step.

\begin{algorithm*}[!t]
\refstepcounter{algorithm}\label{alg:app-whisper-encoder}
\begin{dTwoCodeBox}{Algorithm~\thealgorithm: Stage-I streaming encoder for LLaMA-Omni 2.}
\begin{lstlisting}[style=d2code]
def whisper_step(mel, state, k):
    # mel: [B,128,10*k]; tail: [B,128,2]
    w = cat([state.tail, mel], dim=-1)
    x = gelu(conv2(gelu(conv1(w))))
    x = x.transpose(1, 2)[:, -5 * k :]
    if state.rows.shape[1] == R:
        state.rows = empty_rows()
    state.rows = cat([state.rows, x], dim=1)
    # mask[q,p] = (p // (5*k) <= q // (5*k))
    h = transformer(state.rows + positions(), mask=block_causal(width=5 * k))
    h = norm(h[:, -5 * k :]).reshape(B, k, 6400)
    state.tail = mel[..., -2:]
    return linear2(relu(linear1(h))), state
\end{lstlisting}
\end{dTwoCodeBox}
\end{algorithm*}

\subsection{Qwen3-Omni-D2 Training and Inference}

Let $T$ be the number of 80-ms encoder frames in a training segment,
with $T$ divisible by $k$. User and self embeddings have shape
$[B,T,2048]$, text targets $[B,T]$, and codec targets $[B,T,16]$.
For each chunk, we serialize
\[
 [\,U_{i,1:k};\ A_{i,1:k};\ e(y_{i-1,k});\
 e(y_{i,1:k-1})\,].
\]
Here, $A_i$ contains the model audio from the preceding chunk.
The resulting Thinker input has shape $[B,3T,2048]$ and uses ordinary
causal attention in this order. Learned stream embeddings distinguish
user and self audio. The first text input is PAD; later inputs use
the preceding target or generated token, including across chunk
boundaries. A task-specific system prompt is prefixed once per
conversation.

\paragraph{Alignment and control.}
Text and codec targets share the 80-ms clock. RESPONSE is placed
eight native frames after the last audible user frame. The first
lexical token and model audio start one frame later; lexical
tokens then occupy consecutive frames, followed by PAD. INTERRUPT
is placed eight native frames after user onset. These offsets are
independent of $k$. Training self audio includes the measured
371-sample playback shift at 16 kHz, followed by the one-chunk
feedback delay. The first self-audio chunk uses a learned silence
embedding.
RESPONSE and INTERRUPT reuse \texttt{<|audio\_start|>} and
\texttt{<|audio\_end|>}; PAD is \texttt{<|duplex\_pad|>}.

\paragraph{Training.}
Algorithm~\ref{alg:app-qwen-training} omits padding masks and
distributed reductions.
\texttt{pack} serializes each chunk in the order above;
\texttt{text\_slots} selects its last $k$ positions. Encoder states
are independent for the two audio streams. The task prompt and
Talker speaker prefix are initialized once in training and inference.

\begin{algorithm*}[!t]
\refstepcounter{algorithm}\label{alg:app-qwen-training}
\begin{dTwoCodeBox}{Algorithm~\thealgorithm: Qwen-D2 training on one conversation, with teacher forcing.}
\begin{lstlisting}[style=d2code]
for segment in training_segments:
    U, enc_u = E(segment.user, enc_u)
    A, enc_a = E(segment.self, enc_a)
    A = delay(A, k, previous_self)
    y, c = segment.text, segment.codec
    z = shift_right(y, previous_text)
    x = pack(U + user_row, A + self_row, embed(z))
    h, h24, kv = Thinker(x, kv, causal=True)
    h, h24 = text_slots(h), text_slots(h24)
    logits = text_head(h)  # [B,T,V_text]
    opened = inclusive_BOS_EOS_mask(c[..., 0])
    c_in = shift_right(c, previous_codec)
    c_in[y == RESPONSE] = BOS_frame
    ctx = where(opened, text_proj(embed(y)), hidden_proj(h24))
    v = ctx + where(opened, codec_embed(c_in), 0)
    g, talk_kv = Talker(v, talk_kv, causal=True)
    p0 = codec0_head(g)  # [B,T,V_codec]
    pr = MTP_teacher_force(g, c[..., :15])
    # pr: [B,T,15,V_codec]
    loss = text_CE(logits, y)
    loss = loss + codec_CE(p0, pr, c)
    loss.backward()
    clip_grad_norm_(trainable, 1.0)
    optimizer.step()
    optimizer.zero_grad()
    save_detached_segment_state()
\end{lstlisting}
\end{dTwoCodeBox}
\end{algorithm*}

The text loss uses weights 1 for lexical tokens, 0.05 for PAD, and
20 for RESPONSE and INTERRUPT. Codec targets have weight 1, with
weight 20 for codebook-0 EOS. BOS has no codec loss; residual
codebooks also exclude EOS. The two losses are normalized separately.
We carry detached encoder state, the last $k$ self embeddings, the
last text and codec targets, and about 90 s of Thinker/Talker KV
memory into the next training segment. State is cleared at a
conversation boundary.

\paragraph{Inference.}
Algorithm~\ref{alg:app-qwen-inference} shows the decoding loop.
Each chunk uses the current user audio and the model audio
from the preceding chunk. The model generates and emits speech
one frame at a time, then stores the output as self-audio input
for the next chunk.

\begin{algorithm*}[!t]
\refstepcounter{algorithm}\label{alg:app-qwen-inference}
\begin{dTwoCodeBox}{Algorithm~\thealgorithm: Qwen-D2 inference with delayed self-audio feedback.}
\begin{lstlisting}[style=d2code]
previous_text = PAD
feedback = None
kv = prefill_system_prompt()
state = init_speech_state()
while stream_open:
    user = wait_for_audio(1280 * k)  # 16 kHz
    U, enc_u = E(user, enc_u)
    if feedback is None:
        A = learned_self_silence(k)
    else:
        A, enc_a = E(feedback, enc_a)
    x = cat([U + user_row, A + self_row, embed(previous_text)], dim=1)
    # One prefill of 2k+1 positions
    h, h24, kv = Thinker(x, kv, causal=True)
    output = []
    for j in range(k):
        if j > 0:  # k-1 forwards
            h, h24, kv = Thinker(embed(y), kv)
        logits = text_head(h[:, -1])
        y = sample_allowed(logits, state)
        pcm, state = speech_step(y, h24[:, -1], state)
        output.append(pcm)  # 80 ms, 24 kHz
        emit(pcm)
    previous_text = y
    feedback = resample(cat(output), 24000, 16000)
\end{lstlisting}
\end{dTwoCodeBox}
\end{algorithm*}

In \texttt{speech\_step}, an open Talker conditions on the current
text embedding and preceding codec frame; a closed Talker uses the
Thinker's layer-24 hidden state. RESPONSE opens Talker and supplies
codec BOS. During speech, Talker predicts codebook 0 and MTP predicts
the remaining 15 codebooks autoregressively. INTERRUPT closes an
open Talker with EOS; a sampled codec EOS also ends speech. PAD
ends lexical generation but does not itself close Talker. BOS, EOS,
and closed frames produce silence. Talker KV and Code2Wav state
advance on every native frame, including silence, and rendered
audio is fed back only in the next complete chunk.

\section{Training and Reproducibility Details}
\label{app:training}

\paragraph{Optimization.}
Tables~\ref{tab:app-stage1} and~\ref{tab:app-stage2} summarize the
trainable modules. Full denotes dense fine-tuning of the listed
module. All LoRA adapters use scaling $\alpha=2r$ and zero dropout.
Both stages use AdamW with gradient clipping at 1.0. Computation uses
BF16 with FP32 trainable parameters and optimizer states. Frozen
Qwen Thinker/Talker routed experts use weight-only INT8 storage
during Stage II and are dequantized for computation.

\begin{table}[t]
\centering
\tablefont
\begin{tabular}{llr}
\toprule
Module (both backbones) & Training & LR \\
\midrule
Encoder frontend & Full & $5\times10^{-5}$ \\
Encoder attention & Full & $2\times10^{-6}$ \\
Encoder MLP/projector & LoRA, $r=32$ & $10^{-4}$ \\
Original teacher & Frozen & -- \\
\midrule
\multicolumn{2}{l}{Updates} & 30,000 \\
\multicolumn{2}{l}{Global batch / accumulation} & 4 / 1 \\
\multicolumn{2}{l}{Weight decay} & 0.01 \\
\multicolumn{2}{l}{Learning-rate schedule} & Constant \\
\bottomrule
\end{tabular}
\caption{Stage-I training. Each chunk size has its own
distilled encoder. Hardware and logged time are in
Table~\ref{tab:app-training-time}.}
\label{tab:app-stage1}
\end{table}

\begin{table*}[!t]
\centering
\tablefont
\setlength{\tabcolsep}{4pt}
\begin{tabularx}{\textwidth}{@{}lXlr@{}}
\toprule
Backbone & Module & Training & Peak LR \\
\midrule
Both & Thinker attention & LoRA, $r=128$ & $5\times10^{-5}$ \\
Both & Encoder frontend & Full & $2\times10^{-6}$ \\
Both & Encoder attention/MLP/projector & LoRA, $r=32$ & $10^{-5}$ \\
Both & New control and stream embeddings & Full & $10^{-4}$ \\
Qwen3-Omni & Initial self-silence embedding & Full & $10^{-4}$ \\
Qwen3-Omni & Talker attention/shared expert, context projections, MTP & LoRA, $r=128$ & $5\times10^{-5}$ \\
Qwen3-Omni & Codec embeddings & Full & $5\times10^{-5}$ \\
Qwen3-Omni & Codec output heads & Full & $10^{-4}$ \\
Qwen3-Omni & Small pretrained interfaces (e.g., Talker norms) & Full & $2\times10^{-5}$ \\
LLaMA-Omni 2 & Native TTS attention and MLP & LoRA, $r=128$ & $5\times10^{-5}$ \\
\midrule
Both & Updates / global batch / accumulation & \multicolumn{2}{l}{10,000 / 4 / 1} \\
Both & Weight decay / warmup & \multicolumn{2}{l}{0 / 200 updates} \\
Both & Training segment / detached memory & \multicolumn{2}{l}{90 s / 90 s, rounded down to whole chunks} \\
\bottomrule
\end{tabularx}
\caption{Stage-II training. Unlisted pretrained parameters are frozen,
including Qwen routed experts and routers, Code2Wav, and the LLaMA-Omni 2
TTS input bridge, tokenizer, flow model, and vocoder.}
\label{tab:app-stage2}
\end{table*}

\paragraph{Sampling and segmentation.}
Stage I alternates four LibriSpeech batches with one AVQA batch.
Stage II cycles through two InstructS2S cohorts, one SODA cohort,
and sixteen AVQA cohorts. Each cohort contains four examples.
Long conversations contribute several training segments, while short AVQA
examples usually contribute one. Each training segment produces one optimizer
update; gradients do not accumulate across segments. Four-GPU runs
use one example per GPU, and two-GPU runs use two examples per GPU.
Training and augmentation use seeds 1337 and 13371, respectively.

\paragraph{Initialization and schedules.}
Stage II starts from the Stage-I encoder trained at the same $k$. Encoder MLP
adapters are retained; fresh attention adapters are added over its
now-frozen dense attention. Qwen learning rates decay after warmup
to their floors at update 2,000 and remain there. Thinker/Talker/
context/MTP LoRA uses an absolute floor of $10^{-5}$; other groups
use 10\% of peak. LLaMA-Omni 2 uses cosine decay to 10\% of peak
at update 10,000. Its audio EOS loss weight is 3; other text and
ordinary audio weights follow the main text.

\begin{table*}[!t]
\centering
\tablefont
\begin{tabular}{lrrrrrr}
\toprule
Backbone & $k$ & \thead{Chunk\\duration (ms)} & Stage-I GPUs & Stage-I time (h) & Stage-II GPUs & Stage-II time (h) \\
\midrule
Qwen3-Omni & 1  &   80 & $4\times$ RTX & 1.73 & $4\times$ RTX & 12.5 \\
           & 2  &  160 & $2\times$ L40S & 1.81 & $4\times$ RTX & 13.2 \\
           & 4  &  320 & $2\times$ L40S & 1.58 & $2\times$ H200 & 13.0 \\
           & 8  &  640 & $2\times$ L40S & 2.04 & $4\times$ RTX & 11.9 \\
           & 13 & 1040 & $4\times$ RTX & 1.21 & $4\times$ RTX & 10.8 \\
\midrule
LLaMA-Omni 2 & 1 & 100 & $4\times$ RTX & 1.50 & $4\times$ RTX & 12.4 \\
             & 2 & 200 & $4\times$ RTX & 1.49 & $4\times$ RTX & 11.1 \\
             & 4 & 400 & $4\times$ RTX & 2.09 & $4\times$ RTX & 12.5 \\
             & 8 & 800 & $4\times$ RTX & 2.17 & $4\times$ RTX & 11.2 \\
\bottomrule
\end{tabular}
\caption{Hardware and approximate logged training time. RTX denotes
NVIDIA RTX PRO 6000 Blackwell Server Edition; L40S and H200 denote
NVIDIA L40S and NVIDIA H200. Times are elapsed hours, not GPU-hours, summed within resumed log segments.
They include replayed updates and data/checkpoint overhead, but exclude
queue time and initialization between segments.}
\label{tab:app-training-time}
\end{table*}

\section{Dataset Construction}
\label{app:dataset-construction}

\paragraph{Source data.}
Table~\ref{tab:app-data} lists the prepared source inventory.
Stage I uses only audio; LibriSpeech clips are filtered to 1--20 s
and selected in split and row order. For Stage II, we synthesize
both conversation roles with Kokoro. Each conversation uses one
of 904 user voice recipes and the fixed \texttt{af\_heart} model
voice. We remove exact-zero samples only at utterance edges and
reject conversations containing invalid or all-zero turns. SODA's two alternating
speakers become user and model; an unmatched final user turn
is dropped.

\begin{table}[t]
\centering
\tablefont
\begin{tabular}{llr}
\toprule
\textbf{Source} & \textbf{Unit} & \textbf{Count} \\
\midrule
LibriSpeech & Speech clip & 35,245 \\
AVQA & Audio--QA pair & 35,245 \\
InstructS2S-200K & Conversation & 149,998 \\
SODA & Conversation & 41,776 \\
\bottomrule
\end{tabular}
\caption{Prepared source examples before conversation concatenation
and repeated sampling. Stage I uses only the audio from AVQA.}
\label{tab:app-data}
\end{table}

\paragraph{Full-duplex sequences.}
We preserve turn order and append complete conversations until
sequence duration reaches at least 720 s. Initial silence and
inserted gaps are sampled uniformly from 2--30 s on the frame grid.
At each model-to-user transition, we sample interruption,
direct concatenation, or silence with probabilities 0.5, 0.3,
and 0.2. An interruption starts the next user at a sampled interior
frame of the preceding model utterance. Model speech continues
until the \texttt{INTERRUPT} target. Qwen additionally shortens eligible
user turns with probability 0.1, sampling an audible cutoff in
the latter half; this augmentation is disabled for LLaMA-Omni 2.

\newpage
\paragraph{Labels and augmentation.}
We locate speech boundaries on clean waveforms using two consecutive
10-ms windows above $-60$ dBFS RMS. Control and speech targets are
fixed before acoustic augmentation. Half of the complete user
tracks receive one or two colored-noise components at 8--30 dB
SNR, optional HVAC noise, and sampled gain, filtering, limiting,
clipping, or bit-depth distortion. Augmentation also covers silence
gaps and varies by epoch. Self audio and speech targets stay clean.

\paragraph{AVQA.}
The input concatenates the original audio clip with a spoken
question and its answer choices. The model says the correct
letter followed by the complete choice text. We keep AVQA audio
unaugmented. Source scheduling and training segmentation follow
Appendix~\ref{app:training}.

\section{Speech-Quality Evaluation}
\label{app:speech-quality}

\begin{table*}[!t]
\centering
\tablefont
\begin{tabular}{llr}
\toprule
\thead{Model} & \thead{Chunk size $k$\\(duration)} & \thead{UTMOS $\uparrow$} \\
\midrule
Qwen3-Omni-D2 & 1 (80 ms) & $4.54\pm0.03$ \\
Qwen3-Omni & HD & $4.46\pm0.07$ \\
\midrule
LLaMA-Omni 2-D2 & 1 (100 ms) & $4.16\pm0.20$ \\
LLaMA-Omni 2 & HD & $4.43\pm0.09$ \\
\bottomrule
\end{tabular}
\caption{Speech quality at chunk size $k=1$ for both D2 backbones. UTMOS is reported as mean $\pm$ sample standard deviation on shared prompts. HD denotes the original half-duplex backbone.}
\label{tab:app-utmos}
\end{table*}

\paragraph{Protocol.}
Table~\ref{tab:app-utmos} reports speech quality using UTMOS~\citep{saeki2022utmos}. The evaluation uses the first user turn from 100 InstructS2S rows, excluded from our local D2 training inventory.
All four models receive the same
Kokoro-generated user waveforms and generate their responses without
teacher forcing. D2 checkpoints are taken at update 10,000, with
60 s of silence appended after each prompt for the response.
The original backbones use their released speech-generation paths,
with a 512-token text limit and a 60-s audio limit. Qwen uses the
Chelsie voice in both settings.
The original backbones use BF16 inference.

\paragraph{Scoring.}
We resample each response to 16 kHz and remove only outer silence
using a 20-ms RMS window, a $-50$ dBFS threshold, and 100-ms
margins. Internal pauses and streaming boundaries remain in the
scored waveform. We apply no additional gain normalization.
UTMOS runs in FP32 and assigns one score to each response crop.
The reported mean weights each scorable response equally; the
standard deviation is across responses. Silent outputs and crops
shorter than 0.5 s receive no score. Truncated outputs remain
included.

\paragraph{Results.}
On the 99 mutually scorable Qwen prompts, D2 and its backbone
score 4.54 and 4.46, respectively. All 100 LLaMA-Omni 2 prompts are
scorable; D2 and its backbone score 4.16 and 4.43.

\section{FD-Score Definition}
\label{app:fd_definition}

We define FD-Score in figure~\ref{fig:intro_fdscore} as an equally weighted average of
latency-adjusted turn-taking and interruption take-over
rates (TORs):
\begin{equation}
\mathrm{FD\text{-}Score}
=
\frac{1}{2}
\left(
\frac{R_{\mathrm{TT}}}{1 + L_{\mathrm{TT}}/\tau}
+
\frac{R_{\mathrm{Int}}}{1 + L_{\mathrm{Int}}/\tau}
\right),
\label{eq:fdscore}
\end{equation}
where $R_{\mathrm{TT}}$ and $R_{\mathrm{Int}}$ are the
turn-taking and interruption TORs, expressed as percentage
values on a 0--100 scale. $L_{\mathrm{TT}}$ and
$L_{\mathrm{Int}}$ are their corresponding measured
latencies, and $\tau = 1\,\mathrm{s}$ is a fixed reference
duration. Thus, when latencies are expressed in seconds,
the denominators are numerically $1 + L_{\mathrm{TT}}$
and $1 + L_{\mathrm{Int}}$.

Each term rewards successful interaction while penalizing
latency. For nonnegative latencies, FD-Score lies between
0 and 100; higher scores indicate better performance
under this definition. We compute the two terms separately,
average them, and round only the final score.

For example, Qwen3-Omni-D2 with 80-ms chunks has
$R_{\mathrm{TT}} = 99.2$,
$L_{\mathrm{TT}} = 0.521\,\mathrm{s}$,
$R_{\mathrm{Int}} = 99.5$, and
$L_{\mathrm{Int}} = 0.403\,\mathrm{s}$, yielding
\begin{equation}
\mathrm{FD\text{-}Score}
=
\frac{1}{2}
\left(
\frac{99.2}{1.521}
+
\frac{99.5}{1.403}
\right)
\approx 68.07.
\end{equation}

\begin{comment}
\section{Example Appendix}
\label{sec:appendix}

This is an appendix.
\end{comment}

\end{document}